\documentclass[wcp]{jmlr}

\usepackage{longtable}
\usepackage{comment}
\usepackage{graphicx}
\usepackage{subcaption}
\usepackage{caption}
\usepackage{enumitem}
\usepackage{appendix}
\usepackage{pgfplots}
\usepackage{booktabs}

\makeatletter
\let\Ginclude@graphics\@org@Ginclude@graphics 
\makeatother

\jmlrvolume{}
\jmlryear{2022}
\jmlrworkshop{ACML 2022}

\title[Knowledge Distillation of CNNs]{Knowledge Distillation of Convolutional Neural Networks through Feature Map Transformation using Decision Trees}

 \author{\Name{Maddimsetti Srinivas} \Email{msrinivas@iitkgp.ac.in}\\
  \Name{Debdoot Sheet} \Email{debdoot@ee.iitkgp.ernet.in}\\
  \addr Department of Electrical Engineering, IIT Kharagpur, India }

\editors{}

\begin{document}

\maketitle

\begin{abstract}
The interpretation of reasoning by Deep Neural Networks (DNN) is still challenging due to their perceived black-box nature. Therefore, deploying DNNs in several real-world tasks is restricted by the lack of transparency of these models. We propose a distillation approach by extracting features from the final layer of the convolutional neural network (CNN) to address insights to its reasoning. The feature maps in the final layer of a CNN are transformed into a one-dimensional feature vector using a fully connected layer. Subsequently, the extracted features are used to train a decision tree to achieve the best accuracy under constraints of depth and nodes. We use the medical images of dermaMNIST, octMNIST, and pneumoniaMNIST from the medical MNIST datasets to demonstrate our proposed work. We observed that performance of the decision tree is as good as a CNN with minimum complexity. The results encourage interpreting decisions made by the CNNs using decision trees.    
\end{abstract}

\section{Introduction}
Knowledge distillation \citep{frosst2017distilling} aims to create a lightweight machine learning model termed the student model that is as capable as a computationally expensive teacher model. The student model's success is due to the rich knowledge they leverage from the over-parameterized teacher models. 

Also, the interpretation of decisions made by a neural network is low \citep{dw2019darpa} due to its black-box nature. Meaningful decision-making is essential in many domains like medicine, defense, and law. One of the challenges with designing a neural network-based solution is determining the network's structure, including its width, depth, and operations in the constituent layers. In general, it has been seen that highly accurate models are less interpretable, while highly interpretable models are less accurate. We interpret highly accurate teacher models using explainable student models to achieve the tradeoff between the two paradigms. Apart from having high accuracy, it is also essential to understand the reasoning process of the models.

We replace the neural network part without losing accuracy by taking advantage of decision trees. Our contributions in this paper are 
\begin{itemize}
  \item Extraction of features from the final layer of the CNN using fully connected layers. 
  \item Investigating the feature maps of the CNN using decision trees and comparing the decision tree performance using the extracted features. 
\end{itemize}

CNNs are very popular in the computer vision community. However, a lack of explanation of decision-making by these models restricts them from deploying in real-time applications. The method we introduce in this paper overcomes these limitations. Since the accuracy of the decision trees constructed using extracted features is comparable to the CNN performance, this method can be extended to other layers to explain the decision made by a CNN. The paper is organized as follows. Section 2 discusses the prior art. Section 3 explains our proposed method for feature extraction. The experimental results are presented in Section 4. Finally, Section 5 draws several conclusions.
\section {Prior Art}
Interpretation of neural networks is an application of knowledge distillation. Frosst and Hinton \citep{frosst2017distilling} proposed a method to distill the knowledge from neural networks into a soft decision tree. \cite{liu2018improving} worked on a vanilla decision tree to distill knowledge from deep neural networks with equal depth. Also, Jiawang Bai et al. proposed back-propagation-free rectified decision trees \citep{bai2019rectified} for interpretability of neural networks. Furthermore, Adaptive Neural Trees(ANTs) \citep{tanno2019adaptive}  combine neural networks and decision trees to benefit from both paradigms.
\section{Method}
Our method aims to extract features from the CNN and interpret them using decision trees.
\subsection{CNN Training}\label{CNN}
\begin{figure*}[t]
      \centering
      \centerline{\includegraphics[scale=0.35]{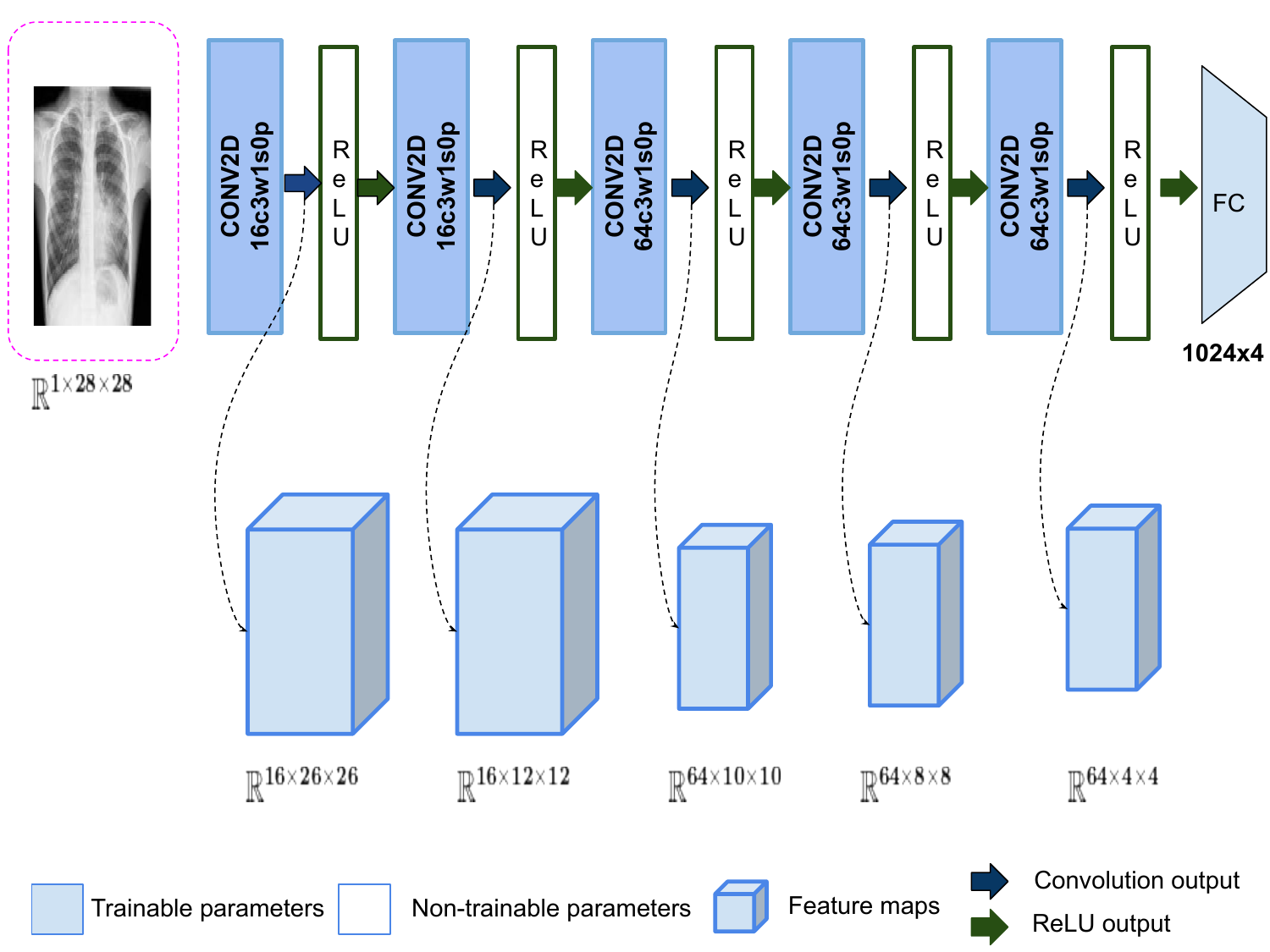}}
    \caption[size=0.5]{CNN Architecture: The kernel size in all the convolutional layers is 3x3, padding is zero, and stride is 1. A fully connected layer constitutes a four-dimensional feature vector in the training phase.}\medskip
    \label{fig:MedNet}
\end{figure*}

Our method mainly depends on the deep learning-based CNN architecture with five convolutional layers, as shown in Figure \ref{fig:MedNet}. Assume that we have $L$ convolutional layers. Let $\mathbf{x}_i\in\mathbb{R}^{C\times H\times W}$ to represent the input image with $H$ rows, $W$ columns, and $C$ color
channels, $\mathbf{y}_i\in\mathbb{R}^{N}$ be the class probability distribution with $N$ classes. Let $\mathbf{z}_i^{kl}\in\mathbb{R}^{C^{l}\times H^{l}\times W^{l}}$ be the feature map for the $l^{th}$ layer of the $i^{th}$ image using $k^{th}$filter, and $\mathbf{d}^{kl}\in\mathbb{R}^{3\times3}$ be the corresponding filter. 

The feature map $z_{i}^{kL}\in\mathbb{R}^{H^{L}\times W^{L}\times C^{L}}$ in the last convolutional layer is turned into a one-dimensional vector using flattening as $\mathbf{u}_i\in\mathbb{R}^{1024}$. The input to the fully connected layer is a $1024$-dimensional vector as the size of the feature map is $4\times4$, and the number of channels is $64$. We use the fully connected layer, which links the $1024$ nodes to four nodes. The number of classes in the dataset decides the number of output nodes. The final probabilities are computed from output nodes using the softmax layer. We compute loss between the ground truth and the predicted class probability using cross-entropy loss as in (\ref{eq:loss}). 
\begin{equation}\label{eq:loss}
    \mathcal{L}_{ce}(y_{c},p_{c}) = -\sum_{c=1}^{N}y_{c} log(p_{c})
\end{equation}
where $\mathcal{L}_{ce}$ is the cross entropy loss, $p_{c}$ is the predictive class probability distribution and $y_{c}$ ground truth distribution.

\subsection{Feature Extraction}

Since we have $64$ channels with the feature map size $4\times4$ connected to four nodes after the flattening operation, as shown in Figure \ref{fig:MedNet}, the mapping function from the feature map to a four-dimensional feature vector can be represented as $\mathfrak{m}:\mathbf{u}_i\rightarrow\mathbf{v}_i\in\mathbb{R}^{1024} \rightarrow \mathbb{R}^{4}$. Therefore, each image sample is transformed into a four-dimensional vector using trained CNN. 

\begin{table}[b]
\centering
\begin{tabular}{cccccc} 
\hline
MNIST dataset & CNN Accuracy & DT Accuracy & Nodes & Leaves & Depth  \\ 
\hline
derma         & 78.5         & 68.1        & 10    & 5      & 4      \\
oct           & 91.5         & 82.4        & 10    & 5      & 4      \\
pneumonia     & 96.5         & 89.8        & 10    & 5      & 4      \\
\hline
\end{tabular}
\caption{Accuracy(\%) comparison of CNN and decision tree (DT) on medical MNIST Datasets. The accuracy, nodes, leaves, and depth are noted for each dataset.}
\label{tab:result}
\end{table}

\begin{figure*}[h]
\centering
  \includegraphics[width=0.7\hsize]{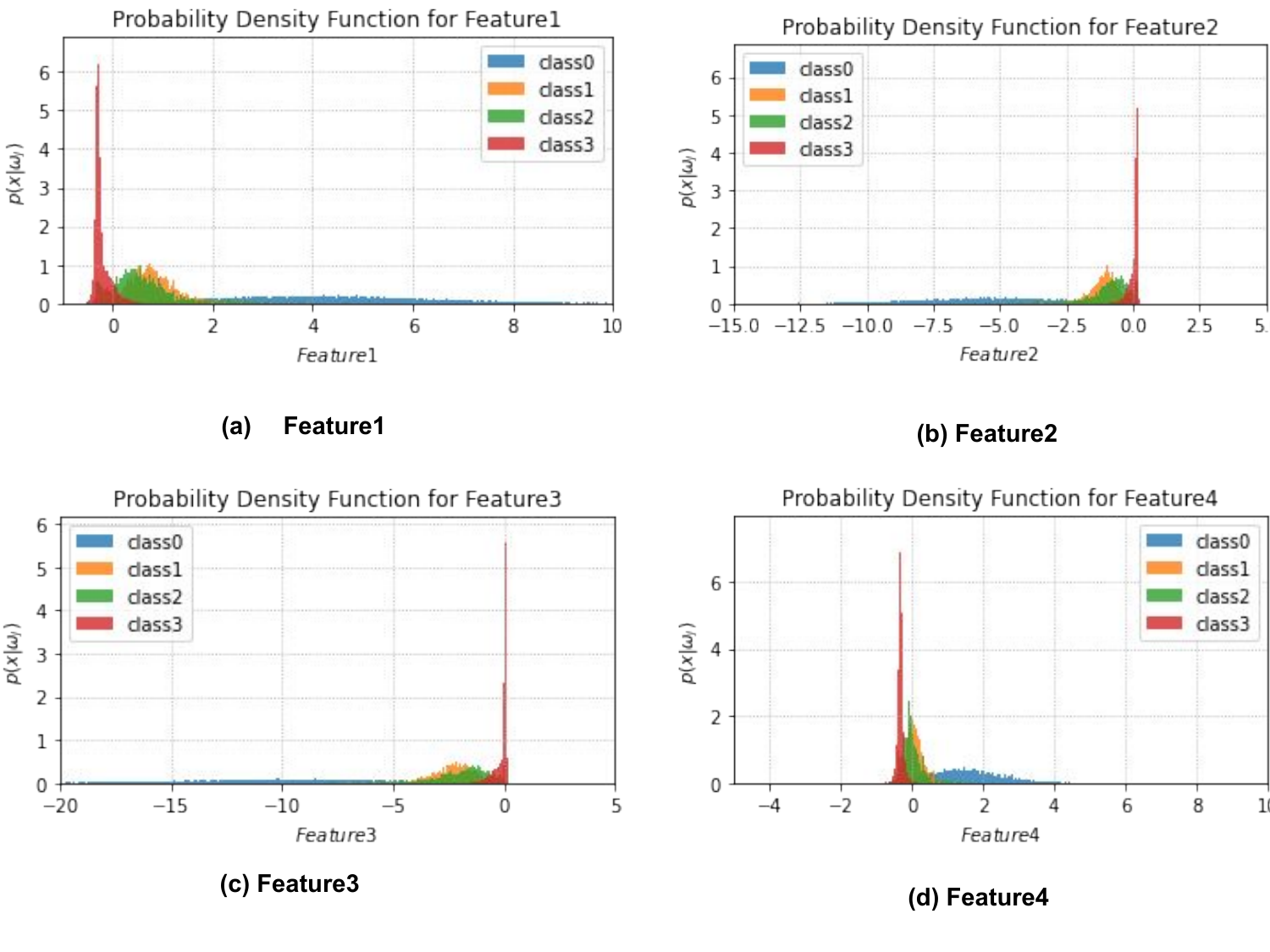}
  \caption{Probability density functions of the four features of $\mathbf{v}_i$. It shows the discrimination capability of $\mathbf{v}_i$ at Layer5 for four classes of the octMNIST dataset.}
\label{fig:pdf}
\end{figure*}
\section{Experiments and Results}
The Medical MNIST v2 datasets \citep{yang2021medmnist} are collected from various sources. We used $70\%$ of the total data samples as training data and $30\%$ of the total data samples as testing data. The hyperparameters are: learning rate is $0.001$, bath size is $128$, epochs are $20$, the optimizer is SGD, momentum is $0.9$, and loss is cross entropy. 

\begin{figure*}[htbp]
    \centering
    \includegraphics[width=0.7\hsize]{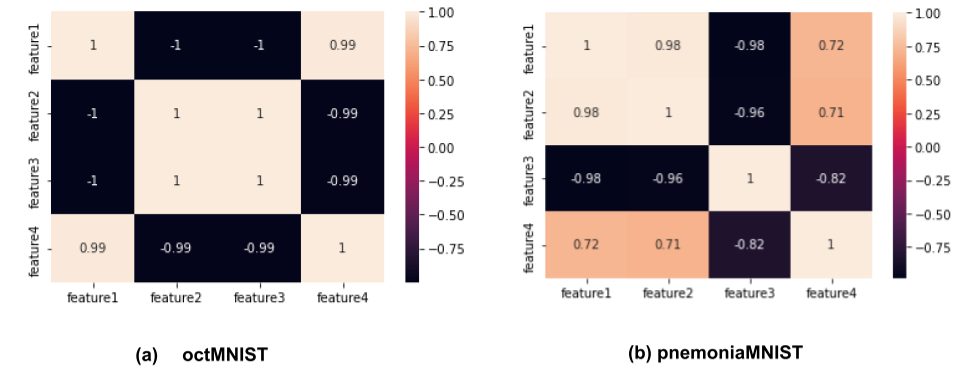}
  \caption{Correlation between the features is shown in the form of $4\times4$ correlation matrix.}
\label{fig:corr}
\end{figure*}

Table \ref{tab:result} compares the accuracy of the CNN with decision trees on the three medical MNIST datasets. The final layer projects the data into a feature space, where the four components of the $\mathbf{v}_i$ obtains high classification ability, as shown in Figure \ref{fig:pdf}. The correlation between the extracted features for the octMNIST dataset and pnemoniaMNIST is shown in Figure \ref{fig:corr}. Since we extract a four-dimensional vector, the size of the correlation matrix is $4\times4$. We observed that \emph {feature1} and \emph {feature4} are highly correlated and negatively correlated with \emph {feature2} and \emph {feature3} for octMNIST. 


\section{Conclusion}
We proposed a method to investigate the features from the final layer of the CNN using decision trees. We transform each image into a four-dimensional feature vector. The acquired data is used as input data to decision trees and investigates the performance of decision trees. Also, we will increase the number of features to observe the discrimination capability of the features and the correlation among the features. Furthermore, we will vary the depth and number of nodes in a tree to improve the performance of the trees. 

\bibliography{acml22}

\end{document}